\documentclass[10pt,twocolumn,letterpaper]{article}

\usepackage{iccv}
\usepackage{times}
\usepackage{epsfig}
\usepackage{graphicx}
\usepackage{amsmath}
\usepackage{amssymb}
\usepackage{textcomp}
\usepackage{subfigure}
\usepackage{enumitem}
\usepackage[accsupp]{axessibility}  

\usepackage[pagebackref=true,breaklinks=true,letterpaper=true,colorlinks,bookmarks=false]{hyperref}

 \iccvfinalcopy 

\def\iccvPaperID{17} 

\ificcvfinal\fi

\begin{document}

\title{Graph-based Neural Architecture Search with Operation Embeddings}

\author{
  \textbf{Michail ~Chatzianastasis}\thanks{Work done as intern at École Polytechnique in Data Science and Mining (DaSciM) team group.} \thanks{Equal Contribution} \\
  National Technical University of Athens \\
  \texttt{mixalisx97@gmail.com}
  \and
  \textbf{George Dasoulas}\footnotemark[2] \\
  École Polytechnique \\
  \texttt{george.dasoulas1@gmail.com}
  \and
  \textbf{Georgios Siolas} \\
  National Technical University of Athens \\
  \texttt{gsiolas@islab.ntua.gr}
  \and
  \textbf{Michalis Vazirgiannis} \\
  Ecole Polytechnique \\ 
  \texttt{mvazirg@lix.polytechnique.fr}

}

\maketitle
\ificcvfinal\thispagestyle{empty}\fi

\begin{abstract}
   Neural Architecture Search (NAS) has recently gained increased attention, as a class of approaches that automatically searches in an input space of network architectures. A crucial part of the NAS pipeline is the encoding of the architecture that consists of the applied computational blocks, namely the operations and the links between them. Most of the existing approaches either fail to capture the structural properties of the architectures or use hand-engineered vector to encode the operator information. In this paper, we propose the replacement of fixed operator encoding with learnable representations in the optimization process. This approach, which effectively captures the relations of different operations, leads to smoother and more accurate representations of the architectures and consequently to improved performance of the end task. Our extensive evaluation in ENAS benchmark demonstrates the effectiveness of the proposed operation embeddings to the generation of highly accurate models, achieving state-of-the-art performance. Finally, our method produces top-performing architectures that share similar operation and graph patterns, highlighting a strong correlation between the structural properties of the architecture and its performance.

\end{abstract}

\section{Introduction}\label{sec:intro}

Deep learning has been in the middle of a research outburst during the last years and the constant need for highly accurate models requires extensive architecture engineering. Neural Architecture Search (NAS) has emerged as the most promising field for the efficient automated search and generation of state-of-the-art models. Its contribution has been studied for a variety of tasks, ranging from medical image segmentation~\cite{medical_image} to objection detection~\cite{object_detection} and speech recognition~\cite{speech_recognition}.

The research interest has focused on latent space optimization techniques~\cite{pmlr-v80-bojanowski18a}, due to their efficiency with respect to the search space and the optimization \cite{yan2020does}. Specifically, a generative model learns continuous representations of neural network architectures, and then the objective function (i.e., the performance of the architecture) is optimized on the latent space. Recent work has shown that the representation of the architecture is crucial for the overall performance of the NAS method \cite{white2020study,bananas}.
The most promising approaches represent the architecture as a graph, in which every node is associated with a layer operation. 
However they assume a fixed encoding of the operations, such as one-hot vectors~\cite{zhang2019dvae}. This assumption puts a limitation on both the expressivity of the operation information and the possible relations between the operations, as it employs orthogonal representations with equal distances between them. 

In this work, we suggest the replacement of the fixed representation of the operations with learnable embeddings that are integrated as parameters into the optimization. Our goal is to produce more accurate and smooth architecture representations, that can take into account how the different operators interact with each other. Our contributions can be summarized as follows:
\begin{itemize}
    \item We propose the operation embeddings as a continuous representation of the applied operators and we integrate them as parameters into the NAS pipeline.
    \item We experimentally show that the parameterized representations of the operations lead to the generation of state-of-the-art architectures.
    \item We observe that the top-performing generated architectures share similar structural patterns, with the clustering coefficient and the average path length being strong indicators of the model performance.  
    
\end{itemize}


The rest of the paper is structured as follows: In Section~\ref{sec:related_work} we highlight previously developed work in the area of NAS. In particular, we emphasize on the application of graph learning algorithms for the encoding of the neural network architecture. In Section~\ref{sec:method} we present our main contribution, which is the introduction of \textit{operation embeddings} into Graph VAEs. Finally, in Section~\ref{sec:exps} we evaluate the contribution of the operation embeddings to VAE models of the neural network architectures through an experimental study in the ENAS search space. Moreover, we investigate how several structural characteristics of the network encoding affect the performance of the generated architectures. 

\section{Related Work} 
\label{sec:related_work}

\paragraph{Neural Architecture Search.}
In the last years, significant progress  has been made in automating architecture engineering. Neural Architecture Search has proved its ability to construct architectures that achieve state-of-the-art results in various tasks, with little human intervention \cite{zoph2017neural,zoph2018learning,runge2018learning,real2019regularized,ghiasi2019nasfpn}. The NAS task can be formulated as an optimization problem in an input space of network architectures. Common techniques for solving this optimization problem as reinforcement learning~\cite{baker2017accelerating,zoph2017neural} and evolutionary methods \cite{real2017largescale,real2019regularized,suganuma2017genetic} operate in a discrete search space. Directly searching an architecture within this space is inefficient given its exponential growth as the number of operations and layers increases \cite{luo2019neural,elsken2018neural}. To tackle this challenge, recent research works have introduced differentiable search methods, that operate on a continuous relaxation of the search space \cite{luo2019neural,liu2018darts}. In particular, Neural Architecture Optimization (NAO) has been proposed as a framework that trains an auto-encoder and a performance predictor using gradient descent~\cite{luo2019neural}. However, the model is trained in a supervised manner, limiting the ability to transfer the latent space in other datasets. In this work, we are showing how our method can be applied to unsupervised NAS models.

\paragraph{Graph Representation Learning for Neural Architecture Search.}
	Unsupervised graph representation learning methods have shown promising results in neural architecture search and specifically in the accurate and expressive architecture encoding~\cite{li2020neural,yan2020does,zhang2019dvae}. The basic idea is to represent a neural network architecture as a graph and to learn a smooth continuous latent space, such that high-performance architectures be mapped close to each other. Given a continuous and smooth architecture representation, various strategies can be efficiently applied, such as the bayesian optimization~\cite{k2018neural,bananas}.
	
    Modeling of network architecture as a graph can be achieved in various ways. String-based methods have been proposed before in order to provide representations of the architecture as a graph ~\cite{bowman2016generating,you2018graphrnn}. These methods represent the graph as a sequence of strings and apply Recurrent Neural Network (RNN) models to process the sequence. The disadvantage of these approaches is that they do not preserve the permutation invariance property~\cite{deepsets, xu2019powerful}, imposing restrictions to the expressiveness of the representations. 
    
    In contrast to the string-based methods, recent research works leverage the structure of the graph and operate directly on it, using message-passing operations~\cite{zhang2019dvae, li2020neural}. D-VAE proposed as a graph-based autoencoder for Directed Acyclic Graphs (DAGs)~\cite{zhang2019dvae}. It applies a graph neural network model with an asynchronous message passing process to encode the architectures. The main limitation of this approach is the utilization of a fixed one-hot encoding of the operation blocks, not being able to capture possible operation relations. Variational Graph Isomorphism auto-encoder is proposed to obtain unsupervised representations of neural network architectures~\cite{yan2020does}. It leverages Graph Isomorphism Networks (GIN)~\cite{xu2019powerful} to encode the graph architectures into the latent space. However, it decodes the whole graph in one shot and also the operations are represented with fixed uninformative vectors. A recent work proposes the utilization of operation information into a generic graph-based framework for encoding neural network architectures, without applying it, though, to an unsupervised setting of architecture generation~\cite{op_emb_graph}.
    
    Finally, the authors in~\cite{pmlr-v119-you20b} utilize relation graphs for the representation of neural network architectures. They apply network generators in order to construct relation graphs with specific structural characteristics. Our work as well, studies structural properties that affect the performance of neural network architecture, but instead of relation graphs, we use DAGs with operation embeddings for the representation of the network architectures.

\section{Operation Embeddings in Variational Graph Auto-Encoders}
\label{sec:method}

 
In this section we first introduce the necessary notation for: a) modeling neural network architectures as graph structures and b) building various graph generative models. Then, we describe our proposed method that replaces the fixed vector encodings of the operations with learnable representations, called as \textit{operation embeddings}. The proposed embeddings can be easily incorporated into a variety of graph-based models and enhance their performance in neural architecture search tasks.

\subsection{Neural Network Architecture as a Directed Acyclic Graph} 
\begin{figure}[t]
    \centering
    \includegraphics[width=0.4\textwidth]{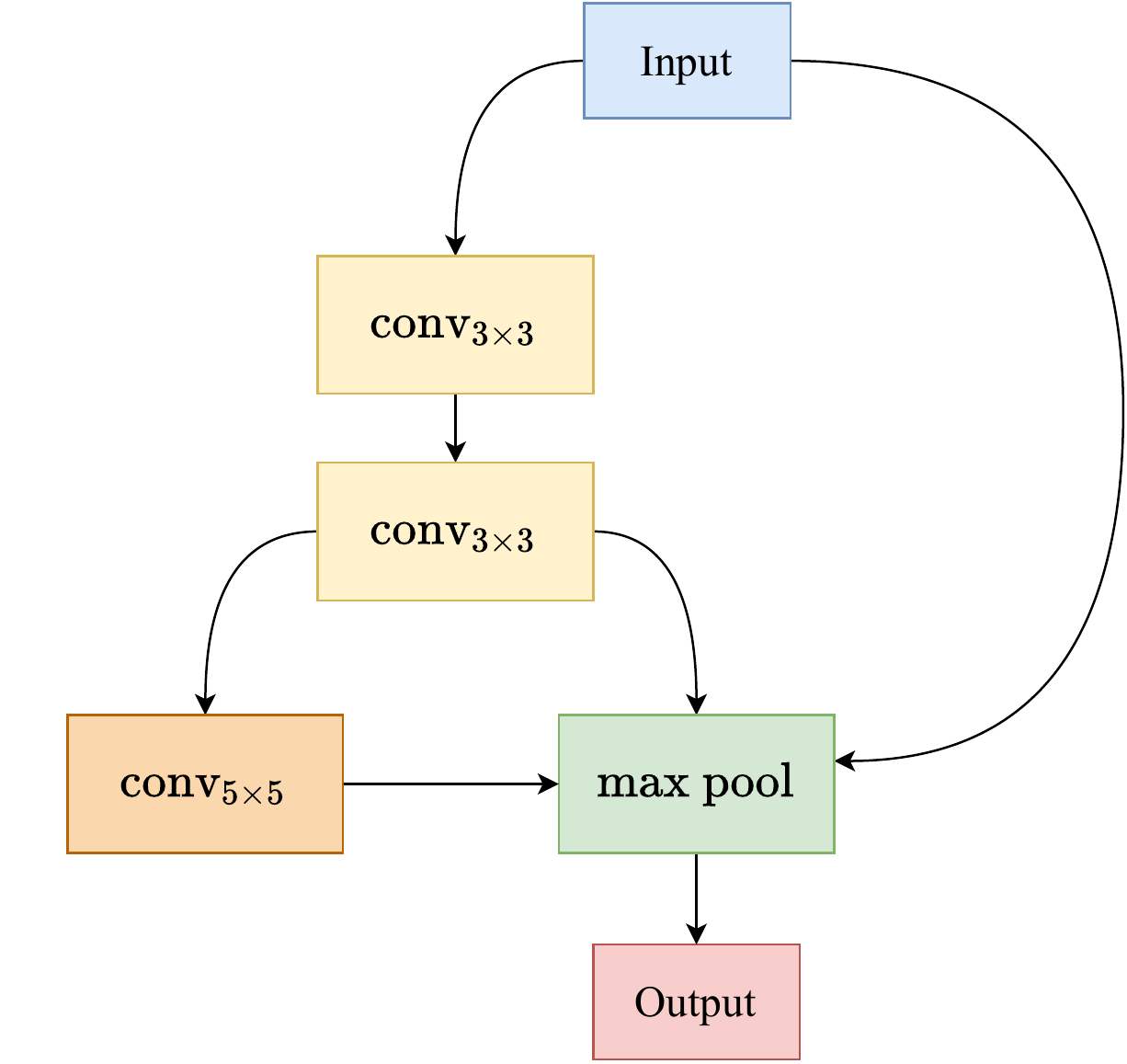}
    \caption{Example of architecture encoding as a computational graph. The used operations are the $3\times 3$ convolution, the $5\times 5$ convolution and the max pooling operator.}
    \label{fig:example_architecture}
\end{figure}

A neural network architecture represents a computation, that is applied to an input signal using a fixed set of operations. We can define the computational graph of an architecture $\mathcal{A}$ as $G_\mathcal{A} = (V,E)$, where $V$ is the set of nodes or the applied operations and $E$ is the set of edges or the links that define the signal flow among the applied operations. We assume that $K$ is the set of the available architecture operations ( e.g an example of K can be $\{\max, \min, \text{conv}_{3\times3}, .. \}$ ). $G_\mathcal{A}$ is a directed, acyclic (i.e., a finite number of performed operations) and labeled graph, where each node $u\in V$ 
is associated with label $x_u$ that corresponds to the operation of node $u$. The most common representation scheme of a labeled graph is the adjacency matrix $\textbf{A} \in \{0,1\}^{|V| \times |V|}$ and the label matrix $\textbf{X} \in \mathbb{Z}^{|V| \times |K|}$, which corresponds to the one-hot encoding of the operations. We note that assuming a DAG structure, $\mathbf{A}$ is not symmetric, imposing an ordering of the nodes and the sequence of processing them. An example of a computational graph of a neural network architecture is visualized in Figure~\ref{fig:example_architecture}.

\subsection{Variational Graph Auto-Encoders}\label{sec:autoencoder}
Let $G=(\textbf{A},\textbf{X})$ be an input graph that represents a neural network architecture.  According
to the standard Variational Graph Auto-Encoder (VGAE) definition~\cite{kingma2014autoencoding}, our goal is to learn a probabilistic encoder model  $q_\phi(\textbf{Z}|\textbf{A},\textbf{X})$ which provides a distribution over latent representations, and a probabilistic decoder model  $p_\theta(\textbf{A},\textbf{X}|\textbf{Z})$ from which we can generate new graphs. We also assume a prior normal distribution over the latent space Z $\sim N(\textbf{0},\textbf{1})$. We train the whole system by minimizing the \textit{evidence lower bound}:
\begin{multline}
\label{eq:lower_bound}
    L(\phi,\theta;\textbf{A},\textbf{X}) = \mathbb{E}_{q_\phi(\textbf{Z}|\textbf{A},\textbf{X})} [\log p_\theta(\textbf{A},\textbf{X}|\textbf{Z})] \\ -\text{KL}[q_\phi(\textbf{Z}|\textbf{A},\textbf{X})||p(\textbf{\textbf{Z}})],
\end{multline} where $\text{KL}$ denotes the Kullback–Leibler divergence. Equation~\ref{eq:lower_bound} indicates that the model does not take into account the performance of an neural architecture and is trained in an unsupervised manner. We make the assumption that architectures with structural similarities and similar operators have similar performance.

\subsubsection{Encoder}\label{sec:encoder}

In the VGAE framework, the encoder uses a graph representation learning model to project $G$ into a representation space with lower dimensionality. More specifically, we use a Graph Neural Network (GNN) model to obtain the representation of the nodes and then we apply a second neural network model to produce the mean and the variance of the posterior approximation. Let a GNN model $ \phi: \mathbb{Z}^{|V| \times |V|} \times \mathbb{Z}^{|V| \times |K|} \rightarrow \mathbb{R}^{|V| \times d}$
denote a graph neural network that takes as input the connections and the operations of the nodes, and outputs an representation of every node. Also, let  $\psi_1, \psi_2: \mathbb{R}^{|V| \times d} \rightarrow \mathbb{R}^l$ denote two differentiable pooling functions that take as input the node representations and output a single representation vector for the whole graph. The encoder can be described via the following equations:
\begin{subequations}
\label{eq:encoder}
\begin{align}
   \mu_G = \psi_1(\phi(\textbf{A},\textbf{X}))), \label{eq:mu_encoder} \\
   \sigma_G = \psi_2(\phi(\textbf{A},\textbf{X}))), \label{eq:sigma_encoder}
\end{align}
\end{subequations}
where $\mu_G$ and $\sigma_G$ denote the mean and the variance of the approximation of the posterior distribution respectively. Note that this formulation expresses multiple variational graph auto-encoder models, that have been used before and utilize either synchronous or asynchronous message-passing processes~\cite{zhang2019dvae,simonovsky2018graphvae,yan2020does}. Moreover, standard choices of $\psi_1, \psi_2$ functions are pooling operators followed by Multi-Layer Perceptrons (MLP).

\subsubsection{Decoder}\label{sec:decoder}
The decoder is responsible for translating the latent representation into graph structures. For this work, we use the autoregressive decoder defined in~\cite{zhang2019dvae}. We, now, briefly describe  the decoder. Given a time step $t$, when node $u_t$ is generated we have the following iterative procedure:
\begin{enumerate}
  \setlength\itemsep{.2em}
    \vspace{-.2cm}

    \item We apply an MLP model, which uses as input the current state of the graph, to determine the type of node $u$.
    \item We update the hidden state of node $u$ using a Gated Recurrent Unit (GRU) model  ~\cite{gru}: $h_{u_t} = gru(\mathbf{x}_{u_t}, h_\text{pred})$, with $h_{\text{pred}}$ denoting the aggregated representation from the predecessors of node $u_t$.
    \item For all time steps $k = t-1, t-2,.., 1$ we apply an MLP model that, given as input the states $h_{u_t}$ and $h_{u_k}$, computes the probability $p_{\text{edge}}$ of existence of edge $(u_k, u_t)$. In case that $p_{\text{edge}} > 0.5$ we add the edge $(u_k, u_t)$ into the DAG and we perform the second step to update the representation $h_{u_t}$.
\end{enumerate}
The iteration stops when the examined node is ending type, and then we output the generated graph structure.

    \begin{figure*}[t]
        \centering
        \includegraphics[width=\textwidth]{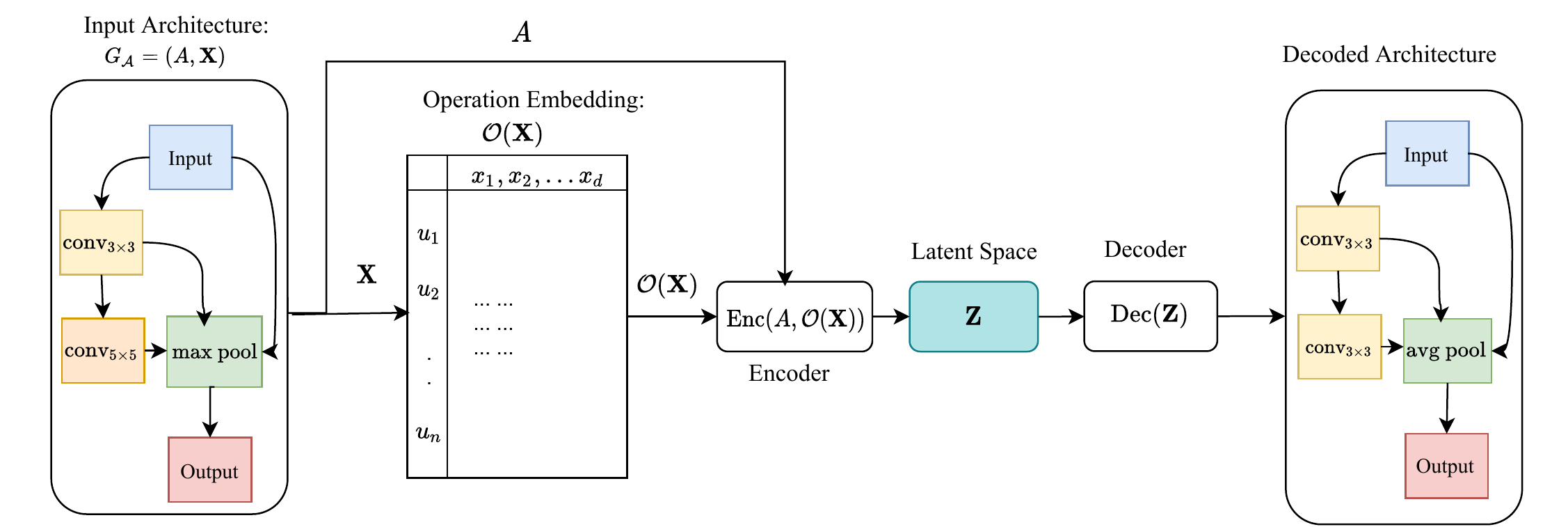}
        \caption{\textbf{Workflow of Variational Graph Auto-Encoder for NAS with integration of Operation Embeddings}. First, the corresponding directed acyclic graph $G_A=(A,\textbf{X})$ of the input architecture is constructed. Then, the operations $X$ are given as input in the operation embeddings layer, to obtain the embeddings $O(\textbf{X})$. Finally, the adjacency matrix and the operation embeddings are passed to the  auto-encoder.
        }
        \label{fig:nas_pipeline}
    \end{figure*}
\subsection{Operation Embeddings}


In the formulation of the models, described in Section \ref{sec:autoencoder}, the operation matrix $\mathbf{X} \in \mathbb{Z}^{V \times |K|}$ is a fixed vector representation. Usually, this representation corresponds to the one-hot encoding of the operation set, so that there is an unordered representation of the operators. Given that, the auto-encoder treats all the operations equally.

The limitations of the one-hot encoding are twofold: 
\begin{enumerate}
  \setlength\itemsep{.2em}
    \vspace{-.2cm}

    \item[a)] It does not take into account the computational relationships and structural dependencies of the different operations. For example, a $5 \times 5$ convolutional layer is more similar to a $3 \times 3$ convolutional layer rather than to a $\text{max-pool}$ layer in terms of the computational level.
    \item[b)] It cannot exploit information from the data, as the one-hot vectors are fixed. This means that the optimization cannot affect the way that the model chooses operations.
\end{enumerate}

Inspired by the success of word embeddings~\cite{NIPS2013_9aa42b31}, we propose the incorporation of the embedding $\mathcal{O}: K \rightarrow \mathbb{R}^{|K|\times d_{op}}$ into the encoder model, to tackle the aforementioned limitations. The mapping $\mathcal{O}(\cdot)$ projects the set of available operations into a $d_{op}$-dimensional continuous space in a differentiable manner. We call the mapping $\mathcal{O}$ \textit{operation embedding}. Equations \ref{eq:mu_encoder} and \ref{eq:sigma_encoder} are transformed as follows:
\begin{subequations}
\label{eq:encoder_emb}
\begin{align}
   \mu_G = \psi_1(\phi(\textbf{A},\mathcal{O}(\textbf{X}))), \label{eq:mu_encoder_emb} \\
   \sigma_G = \psi_2(\phi(\textbf{A},\mathcal{O}(\textbf{X}))). \label{eq:sigma_encoder_emb}
\end{align}
\end{subequations}

 In order to learn the operation embeddings used in equations \ref{eq:mu_encoder_emb} and \ref{eq:sigma_encoder_emb}, we treat them as parameters of the auto-encoder and optimize them with gradient descent along with the other weights. The incorporation of the operation embeddings into the architecture generation pipeline is visualized in Figure~\ref{fig:nas_pipeline}. We note that the same embedding $O(\cdot)$ is shared among the encoder and the decoder.
 
\vspace{-.2cm}
\paragraph{Latent Space}  Our ultimate goal is to produce smooth and accurate latent representations of neural architectures. Essentially, we want architectures with similar performance to be mapped in latent representations that are close to each other. This can help the downstream search algorithm to efficiently discover a distribution of high-performing architectures. Since the parameters of the embeddings matrix are changing throughout the training,
a variable representation, based on the end task, of the operations
is possible. The gradients of $\mathcal{O}(\mathbf{X})$'s weights affect the model training procedure as well. Using the equations \ref{eq:mu_encoder_emb} and \ref{eq:sigma_encoder_emb}, the model is able to map computationally similar operations close to each other. Consequently, architectures with similar structures and operation choices can have similar representations, leading to a smooth latent space. 

\vspace{-.2cm}
\paragraph{Implementation} For this study, we choose low-dimensionality for the produced operation embeddings with $d_{op}=3$, as the number of different operations is small. We fully train the autoencoder model for $N$ epochs and we repeat the process for $T$ iterations. Let $\mathcal{O}_{n,t}(\mathbf{X})$ denote the operation embeddings matrix in $n$-th epoch of the $t$-th iteration.
In the first iteration, we initialize the weights of  $\mathcal{O}_{1,1}(\mathbf{X}
)$ from $\mathcal{N}(0,1)$. In the iteration $t = T_i$, we initialize the operation embeddings using the output of the last epoch in the previous iteration $\mathcal{O}_{N,t-1}(\mathbf{X})$. Using this pre-training schema, we manage to achieve faster convergence of the model among the iterations, as the operation embeddings include more prior knowledge, based on the examined task.

\section{Experiments}\label{sec:exps}
In this section, we empirically evaluate our proposed operation embeddings method. The experimentation details and the code are provided in the supplementary material.

\paragraph{Baselines.} To demonstrate the effectiveness of our approach, we incorporate operation embeddings into two variants of a well known variational graph auto-encoder model for DAGs, that use as encoders either asynchronous message-passing operations (D-VAE)~\cite{zhang2019dvae} or simultaneous graph convolutions (GCN)~\cite{kipf2017semisupervised}. We refer to the models with operation embeddings as DVAE-EMB and GCN-EMB respectively. In DVAE-EMB we repeat the model training for $T=4$ iterations, and in GCN-EMB for $T=1$ as described in Section~\ref{sec:method}. We also include S-VAE \cite{bowman2016generating} and GraphRNN~\cite{you2018graphrnn} as baselines, which represent the architecture as a sequence of strings, and do not operate directly on the graph structure. 

\paragraph{Tasks.} In order to have a fair comparison with other approaches, we follow the experimental setup of \cite{kusner2017grammar,zhang2019dvae}. First, we compute basic effectiveness metrics of the variational auto-encoder models and we measure the \textit{predictive performance} of the latent representations. Next, we present the best-performing architectures obtained with Bayesian optimization on the latent space, and note the observed similarities of several graph characteristics of them. Finally, we visualize the learned latent representations to show their smoothness. 

\paragraph{Dataset.} We train the variational graph auto-encoder models in ENAS search space~\cite{pham2018efficient} for 300 epochs. The dataset contains 19,020 neural architectures. Each architecture has 6 layers besides one input and one output layer. Each layer is associated with one operation. There are six available operations: 3 $\times$ 3 and 5 $\times$ 5 convolutions, 3 $\times$ 3 and 5 $\times$ 5 depthwise-separable convolutions \cite{8099678}, 3 $\times$ 3 max pooling and 3 $\times$ 3 average pooling. We use 90\% of the dataset as training data, and the remaining 10\% for evaluation. For the evaluation of their true performance, we fully train the architectures on CIFAR-10, using the same experimental setup with ~\cite{pham2018efficient}.


\subsection{Basic abilities of Variational Graph Auto-Encoders}

\begin{figure}
	\includegraphics[width=0.40\textwidth]{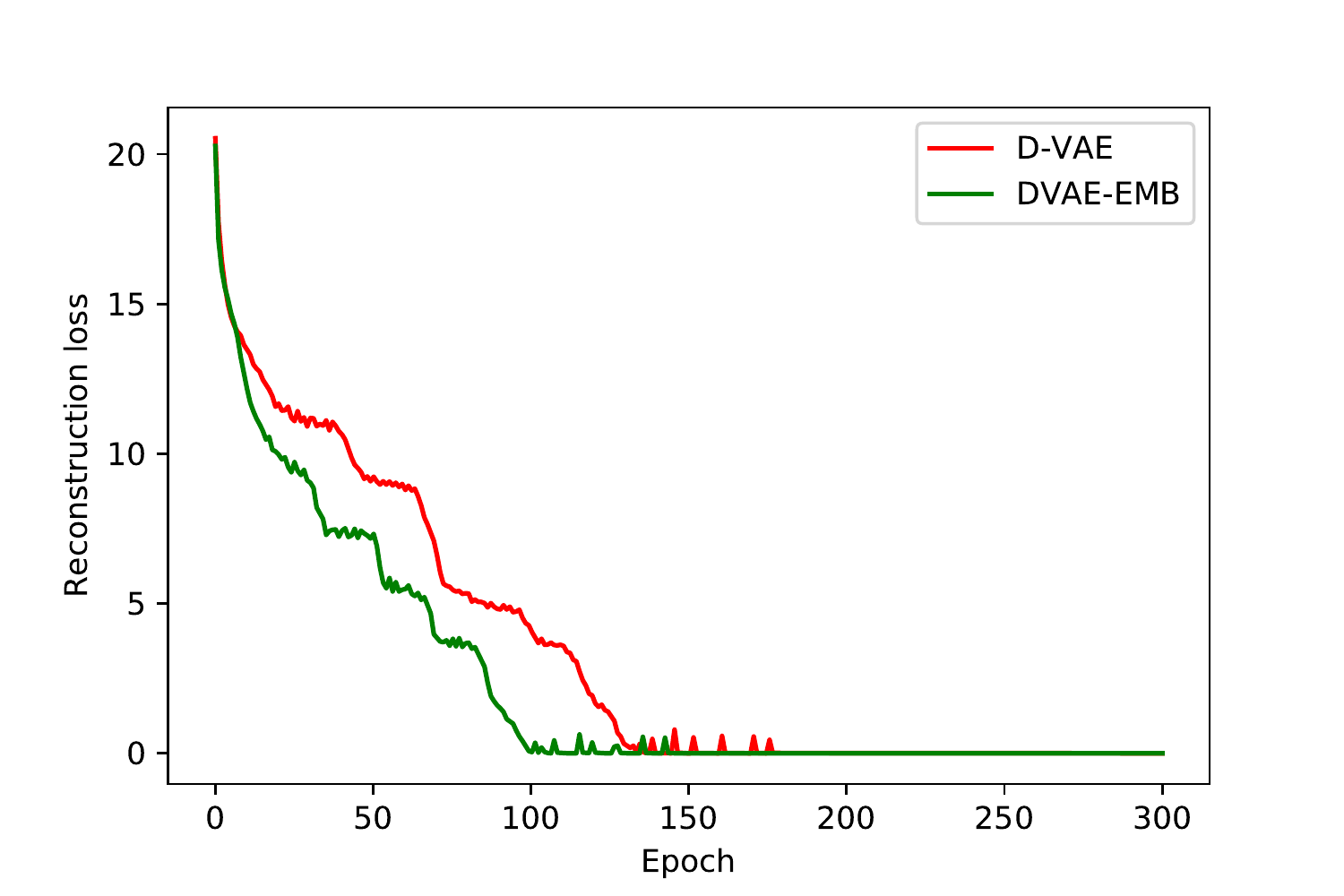}
	\centering
	\caption{The reconstruction loss during the training process.}	
	\label{rec_loss}
\end{figure} 

\begin{figure}
	\includegraphics[width=0.40\textwidth]{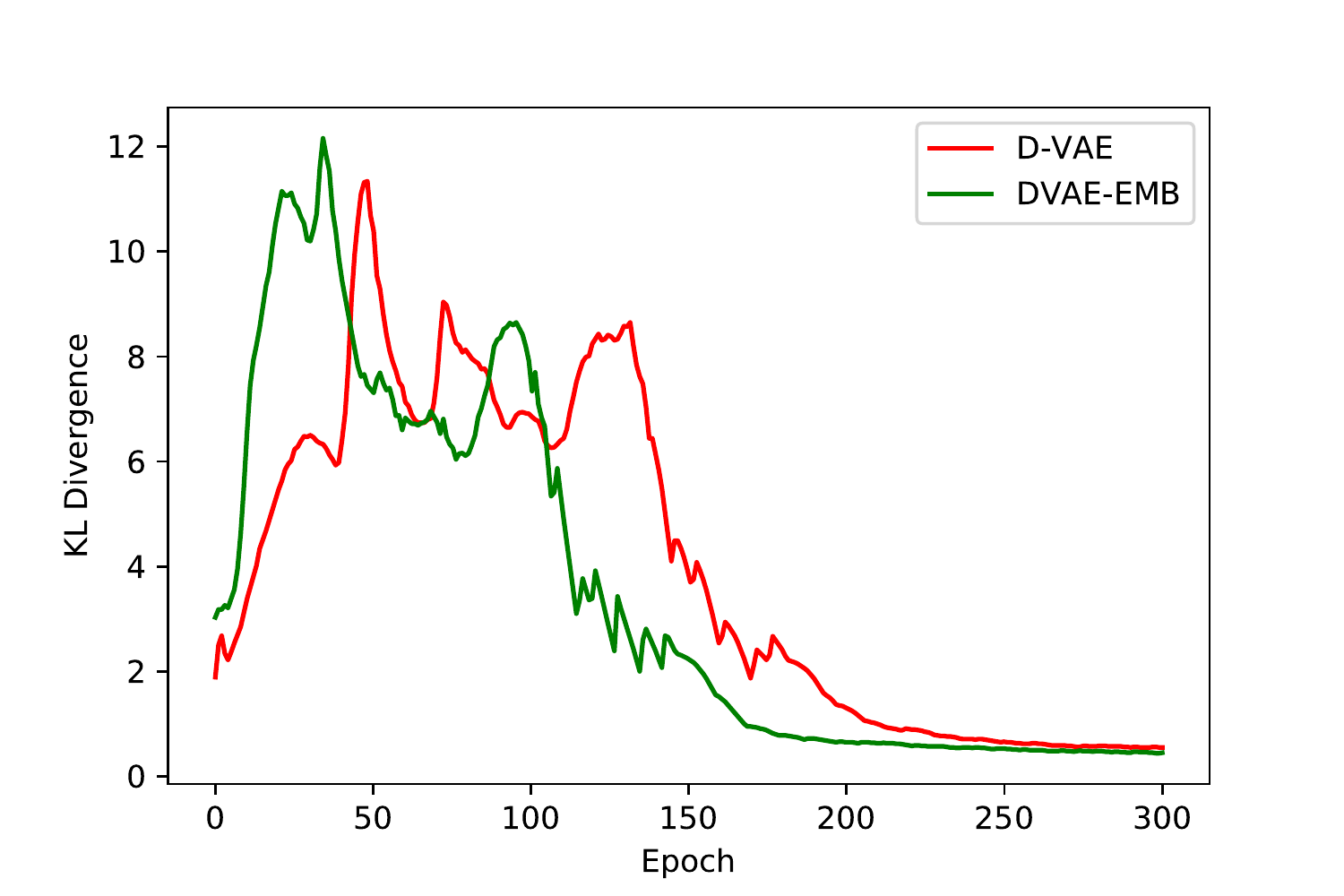}
	\centering
	\caption{The KL divergence during the training process. }	
	\label{kl_loss}
\end{figure} 
In this experiment, we evaluate the reconstructive abilities and the generative properties of the auto-encoders. We use the following metrics proposed by~\cite{zhang2019dvae} : \begin{enumerate}
  \setlength\itemsep{.2em}
    \vspace{-.2cm}
    
    \item \textit{Accuracy}. The percentage of perfectly reconstructed architectures.
    \item \textit{Validity}. The percentage of valid architectures generated from the prior distribution.
    \item \textit{Uniqueness}. The proportion of unique
architectures out of the valid generations.
\end{enumerate}

We present the results in Table \ref{exp1:recon_accuracy}.
DVAE-EMB and GCN-EMB outperform their counterparts in terms of reconstruction accuracy and validity, demonstrating the effectiveness of operation embeddings. D-VAE and GCN have smaller reconstruction accuracy, because their one-hot vector representation fails to capture the operation information accurately. 

We, also, visualize in Figures \ref{rec_loss} and \ref{kl_loss} the reconstruction loss and the KL divergence during the training of D-VAE and our proposed model DVAE-EMB. We observe that the convergence of the reconstruction loss of DVAE-EMB is faster than D-VAE. Intuitively, the incorporation of operation embeddings helps the model to acquire extra information as it captures the relations between the operations. These relations can not be discovered in the D-VAE model, which uses one hot-vectors for representing  the operations. Therefore, our model can converge in fewer epochs achieving lower training loss.

Moreover, in Figure \ref{kl_loss}, we observe a common pattern in the KL-divergence between the two models. In the first epochs, the encoder is quite simple therefore the posterior approximation $q_\phi(z|x)$ is close to the prior $p(z)$. Consequently, the KL divergence has small values. During the optimization of the auto-encoder, the training of the encoder proceeds and the posterior approximation diverges from the prior. As a result, the KL divergence grows. After 100 epochs, when the reconstruction loss is close to zero for each model, the KL divergence starts decreasing because it is the only factor that affect the loss function. 


\subsection{Predictive performance of encoded latent representations}

\begin{table}[t]
    \begin{center}
        \begin{tabular}{|c|c|c|c|}
    \hline
    Model & Accuracy & Validity & Uniqueness \\
    \hline
    \label{exp1:recon_accuracy}
    D-VAE &  99.96 & 100.00 & 37.26  \\
    GCN & 98.70 & 99.53 & 34.00 \\
    S-VAE & 99.98 & 100.00 & 37.03\\
    GraphRNN & 99.85 & 99.84 & 29.77 \\
    \hline
    \textbf{DVAE-EMB} & \textbf{99.99} & \textbf{100.00} & \textbf{39.15} \\ GCN-EMB & 98.87 & 99.95 & 32.63 \\
    \hline
    \end{tabular}
 
    \end{center}
    \caption{Reconstruction accuracy, prior validity and uniqueness results (\%) for the  baselines and our method.}
\end{table}

Next, we evaluate the representation power of the learned latent representations with respect to the performance of the generated neural network architectures. If we can accurately predict the performance based on the latent representations, then we can easily discover the best architectures from the latent space using a downstream strategy. 

Following the experimentation setup in \cite{zhang2019dvae}, we train a Sparse Gaussian Process (SGP) with 500 inducing points on the latent representations of the training data, in order to predict the accuracy of the test architectures. We use two evaluation metrics, the Root Mean Square Error (RMSE) between the Gaussian process predictions and the true performances, and the Pearson correlation coefficient (Pearson's $r$). Pearson correlation coefficient measures the linear correlation between the predictions and the true performances. Therefore, a model with a small RMSE and a high Pearson's $r$ has strong predictive abilities. 
The experiments are repeated 10 times and we report the mean and the standard deviation. 

We show the results in Table \ref{exp2:rmse_pearsons}. The models incorporated with operation embeddings (DVAE-EMB and GCN-EMB) outperform the rest of the models in both metrics. This indicates that the latent spaces of DVAE-EMB and GCN-EMB are more suitable for searching high-performance neural architectures. Comparing DVAE-EMB with GCN-EMB, we observe that DVAE-EMB has the best performance due to its asynchronous message-passing scheme. However, GCN-EMB is significantly better than GCN, an outcome that highlights the contribution of the operation embeddings in learning predictive latent representations. S-VAE and GraphRNN, which leverage neither the graph structure nor the operation embeddings, present low predictive performance.  

\begin{table}[t]
    \begin{center}

        \begin{tabular}{|c|c|c|}
    \hline
    Model & RMSE & Pearson's $r$ \\
    \hline
    \label{exp2:rmse_pearsons}
    D-VAE &  0.384 \textpm 0.002 & 0.920 \textpm 0.001   \\
    GCN & 0.485 \textpm 0.006 & 0.870 \textpm 0.001 \\
    S-VAE &  0.478 \textpm 0.002 & 0.873 \textpm 0.001   \\
    GraphRNN & 0.726 \textpm 0.002 & 0.669 \textpm 0.001   \\
    \hline
    \textbf{DVAE-EMB} & \textbf{0.371 \textpm 0.003} & \textbf{0.925 \textpm 0.001} \\
     GCN-EMB & 0.441 \textpm 0.002 & 0.892 \textpm 0.001 \\
     \hline
    \end{tabular}

    \end{center}
    
     \caption{Predictive performance of encoded latent representations (\%)}
\end{table}

\subsection{Best performing architectures obtained from Bayesian optimization (BO)}
\begin{figure}
\centering  
\subfigure{\includegraphics[width=0.19\linewidth,height=5cm]{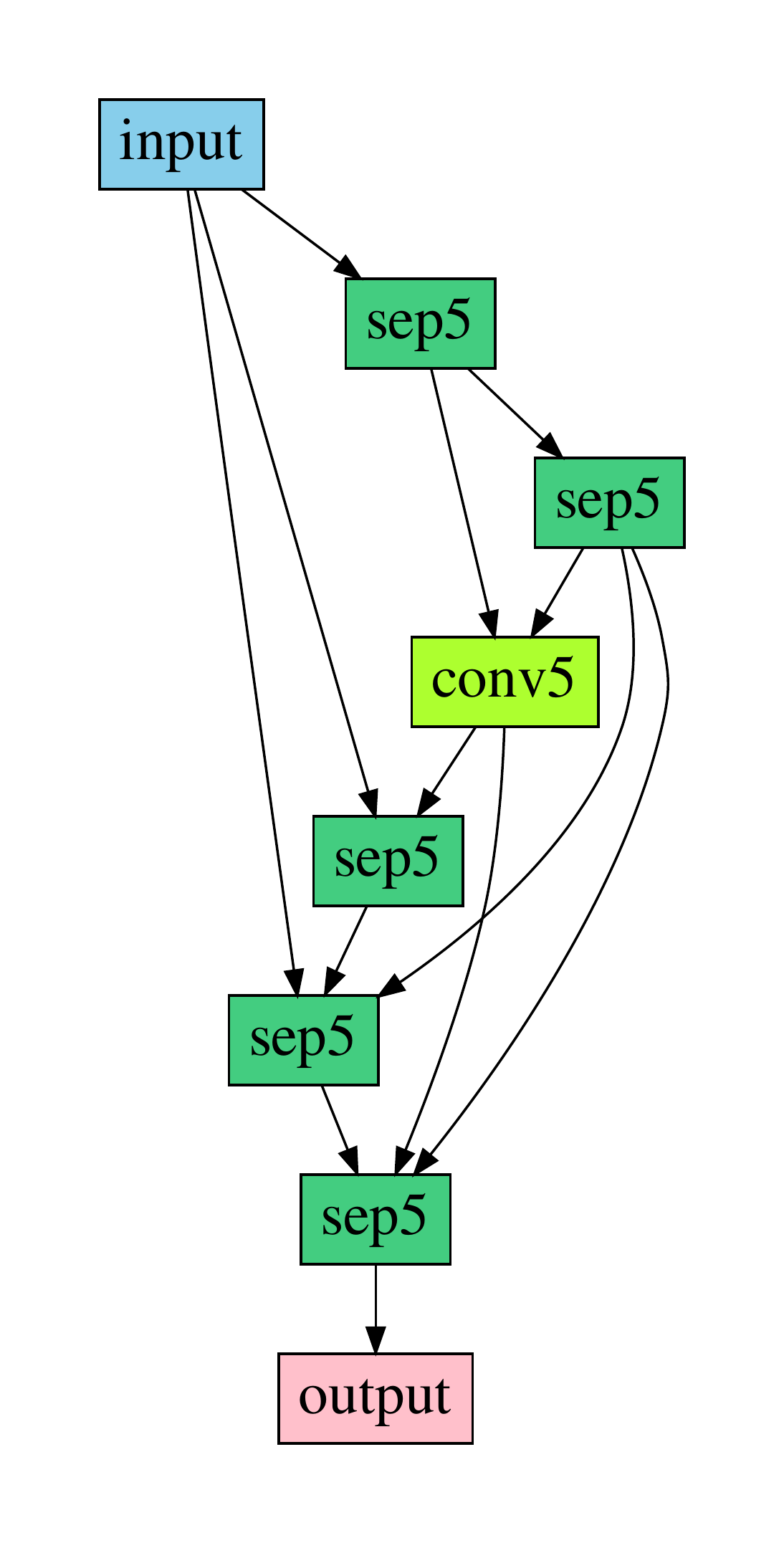}}
\subfigure{\includegraphics[width=0.19\linewidth,height=5cm]{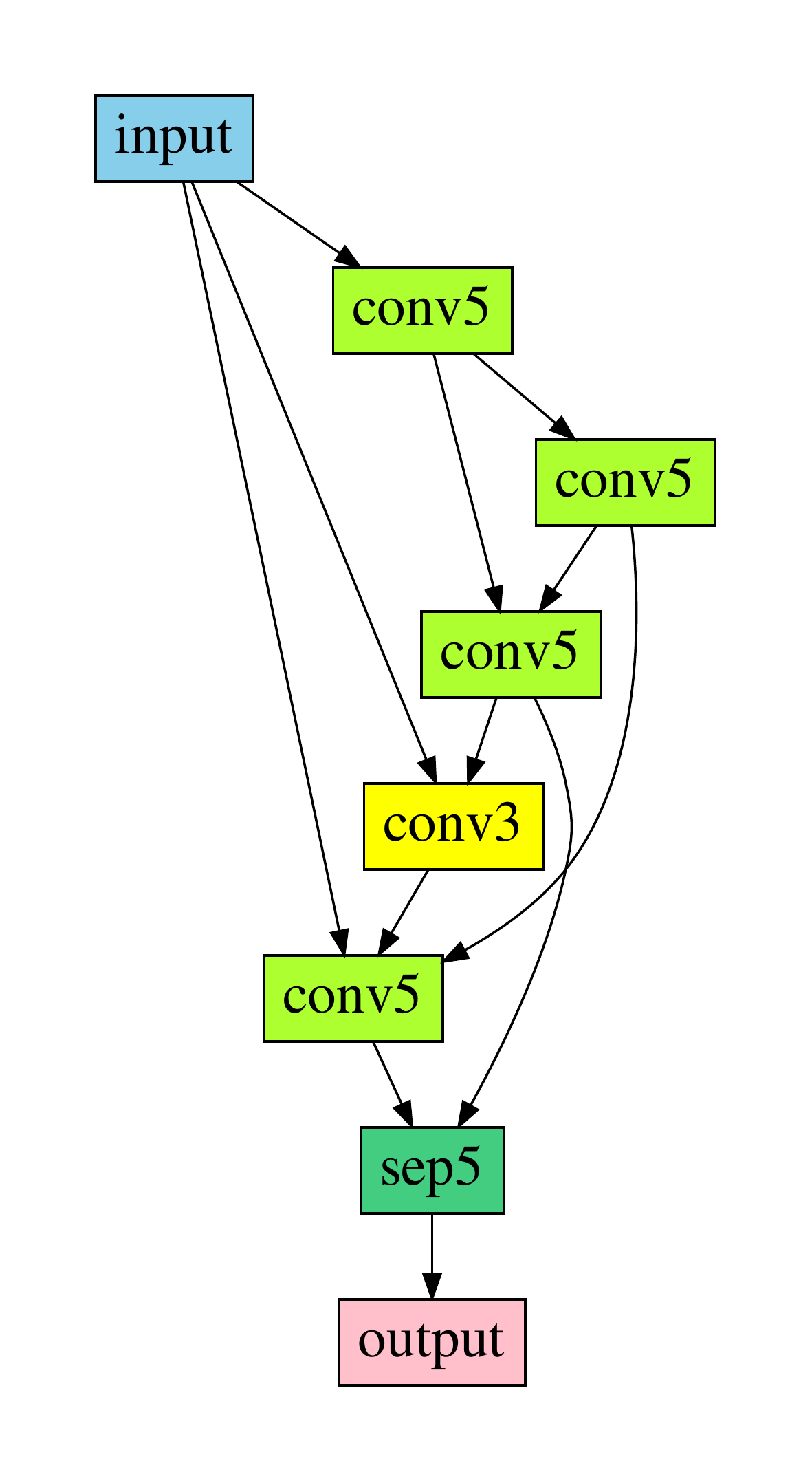}}
\subfigure{\includegraphics[width=0.19\linewidth,height=5cm]{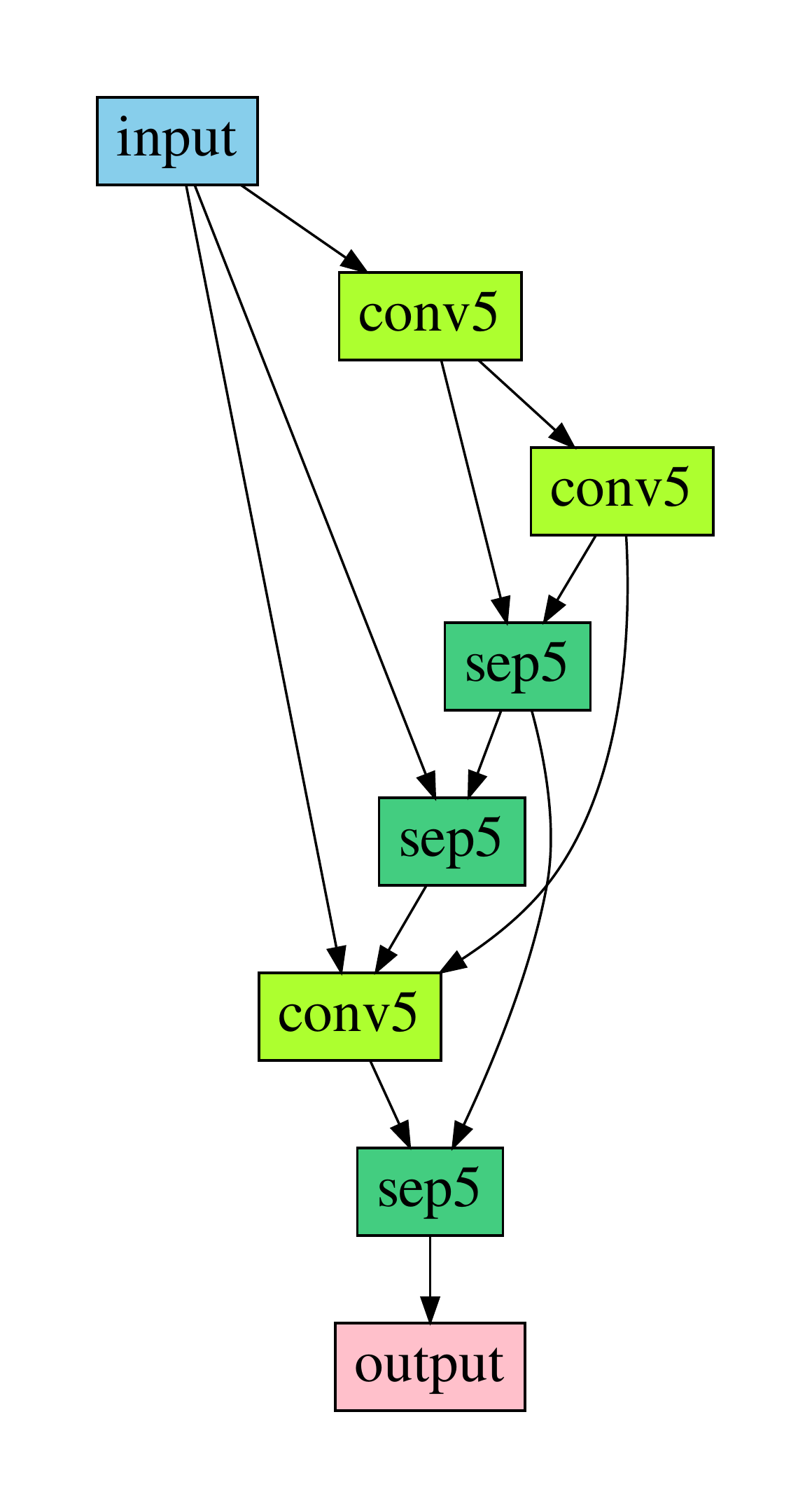}}
\subfigure{\includegraphics[width=0.19\linewidth,height=5cm]{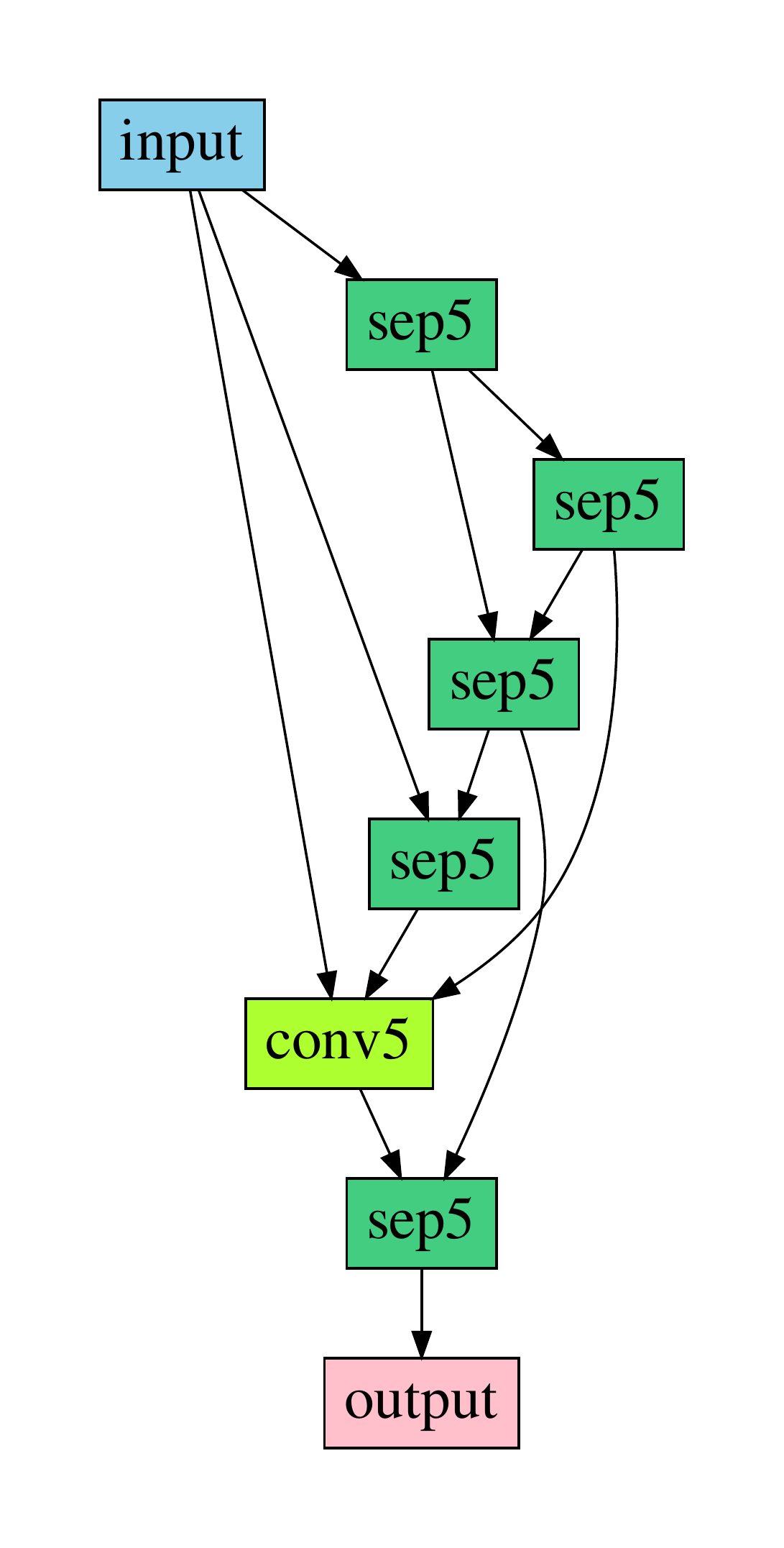}}
\subfigure{\includegraphics[width=0.19\linewidth,height=5cm]{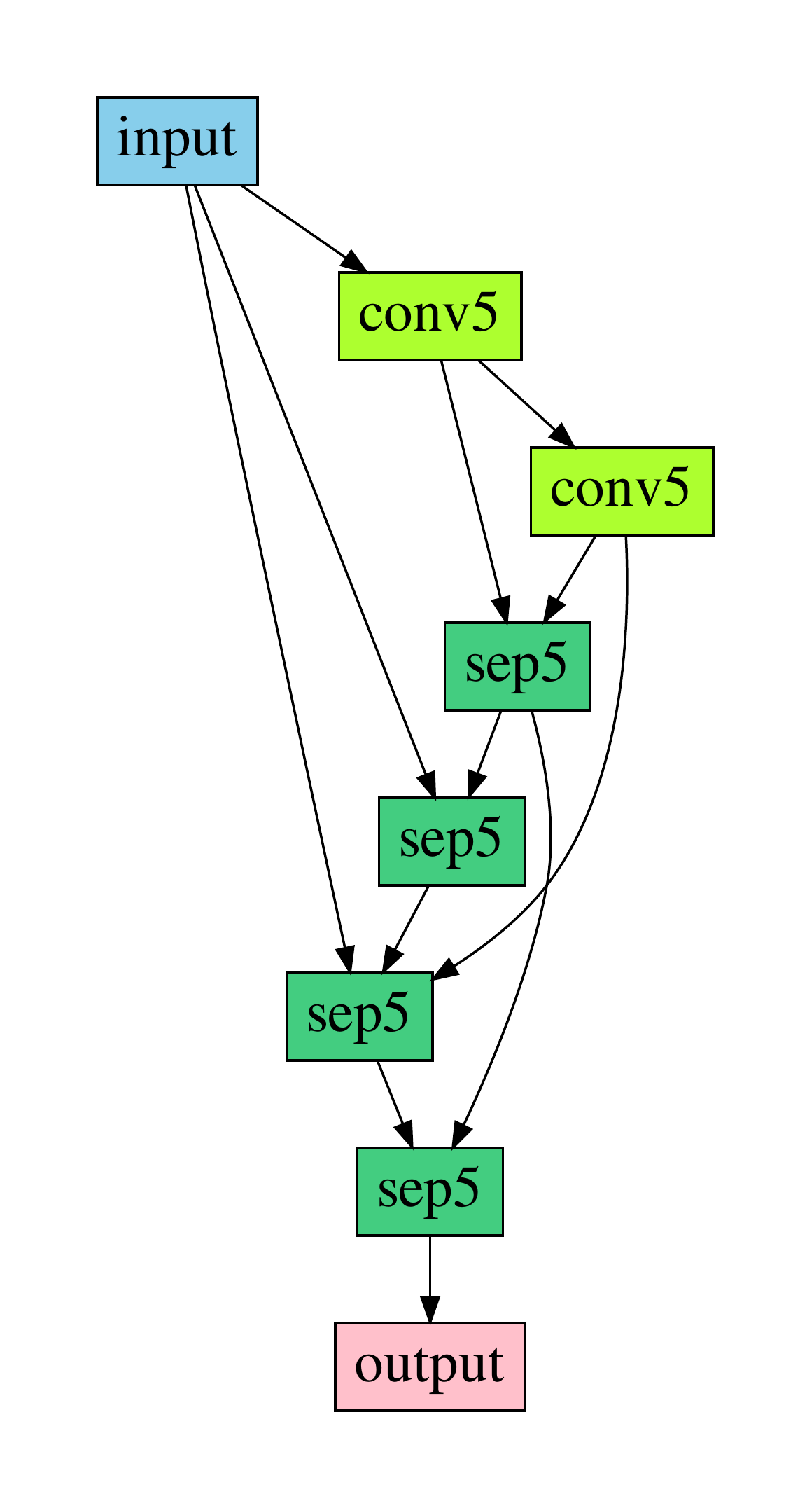}}
\caption{Illustration of the top-5 performing architectures found by \textbf{DVAE-EMB}.The reported test accuracies (from left to right) are: $95.35\% , 95.33\% , 95.17\% , 95.11\%, 95.10$\% .}
\label{exp3:dvae_emb3}
\centering  
\subfigure{\includegraphics[width=0.18\linewidth,height=5cm]{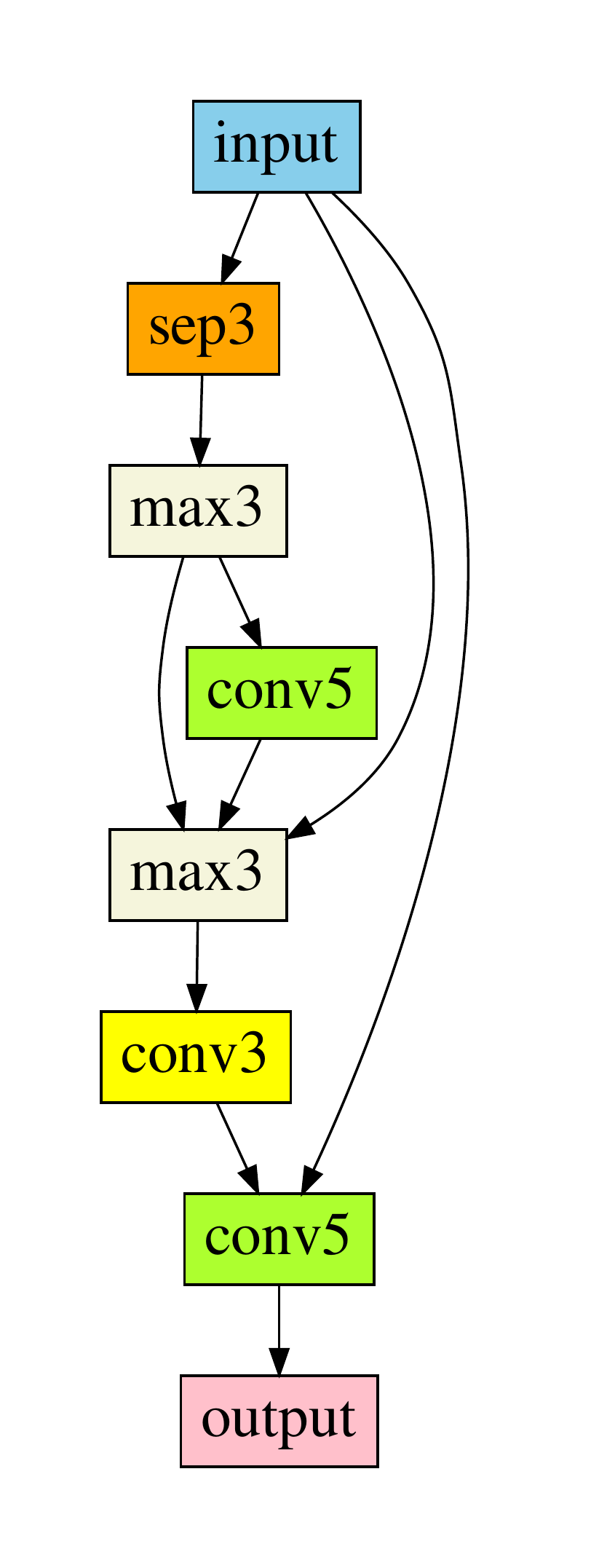}}
\subfigure{\includegraphics[width=0.18\linewidth,height=5cm]{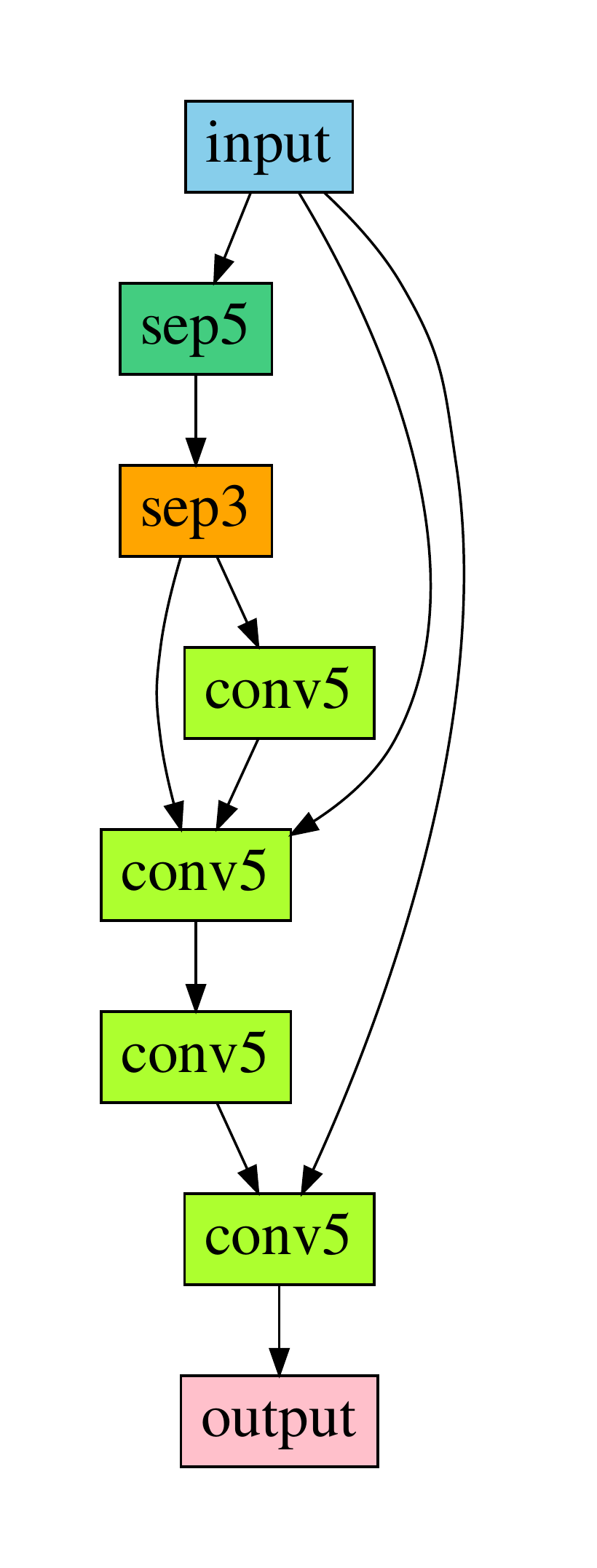}}
\subfigure{\includegraphics[width=0.18\linewidth,height=5cm]{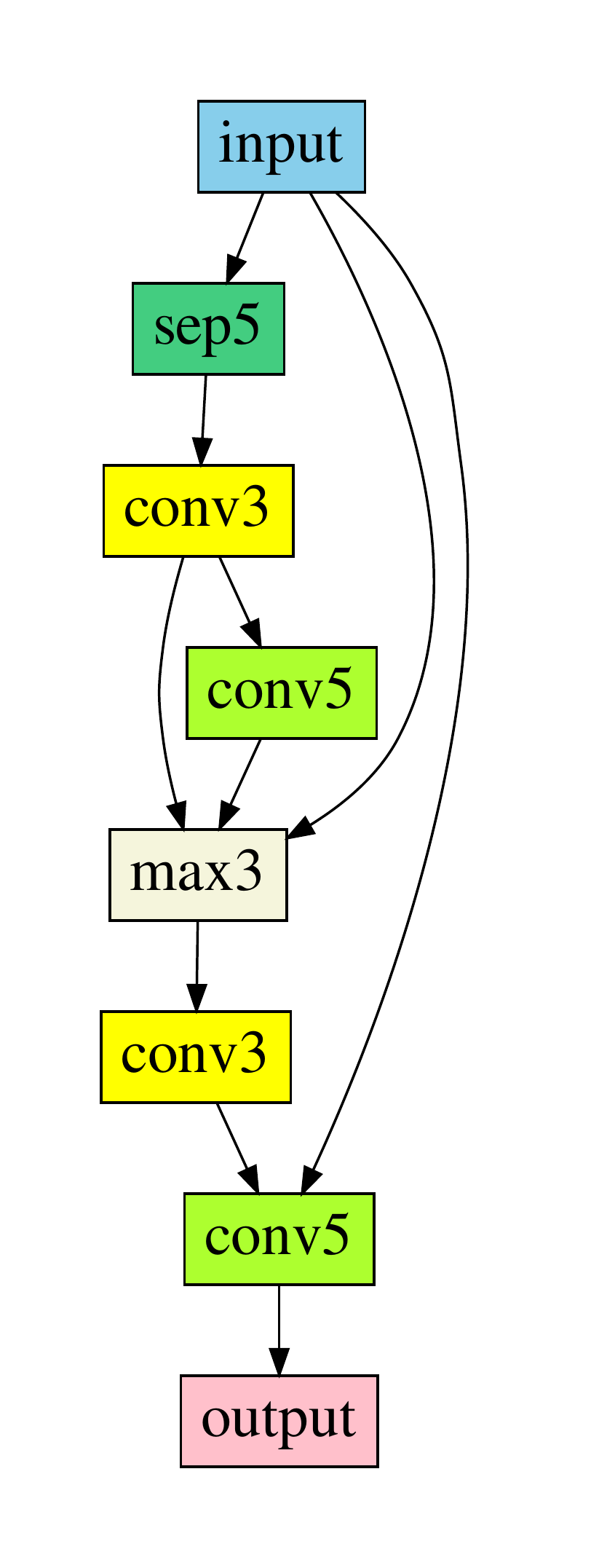}}
\subfigure{\includegraphics[width=0.18\linewidth,height=5cm]{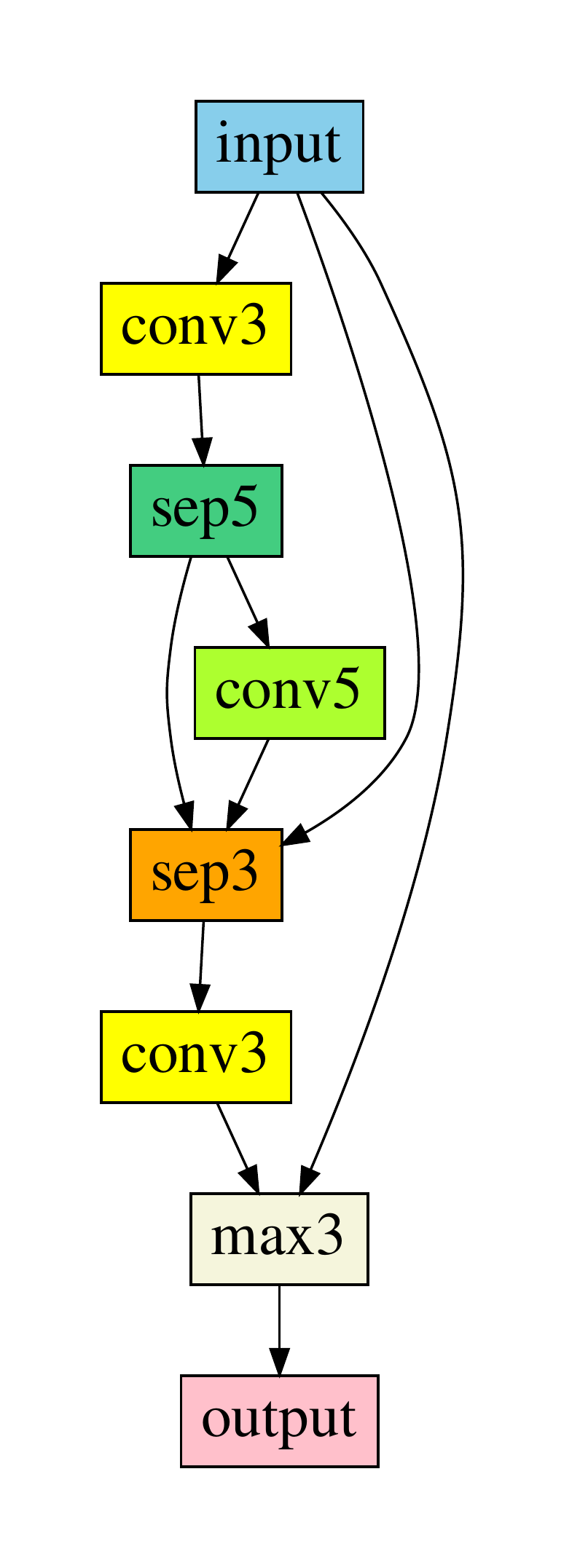}}
\subfigure{\includegraphics[width=0.18\linewidth,height=5cm]{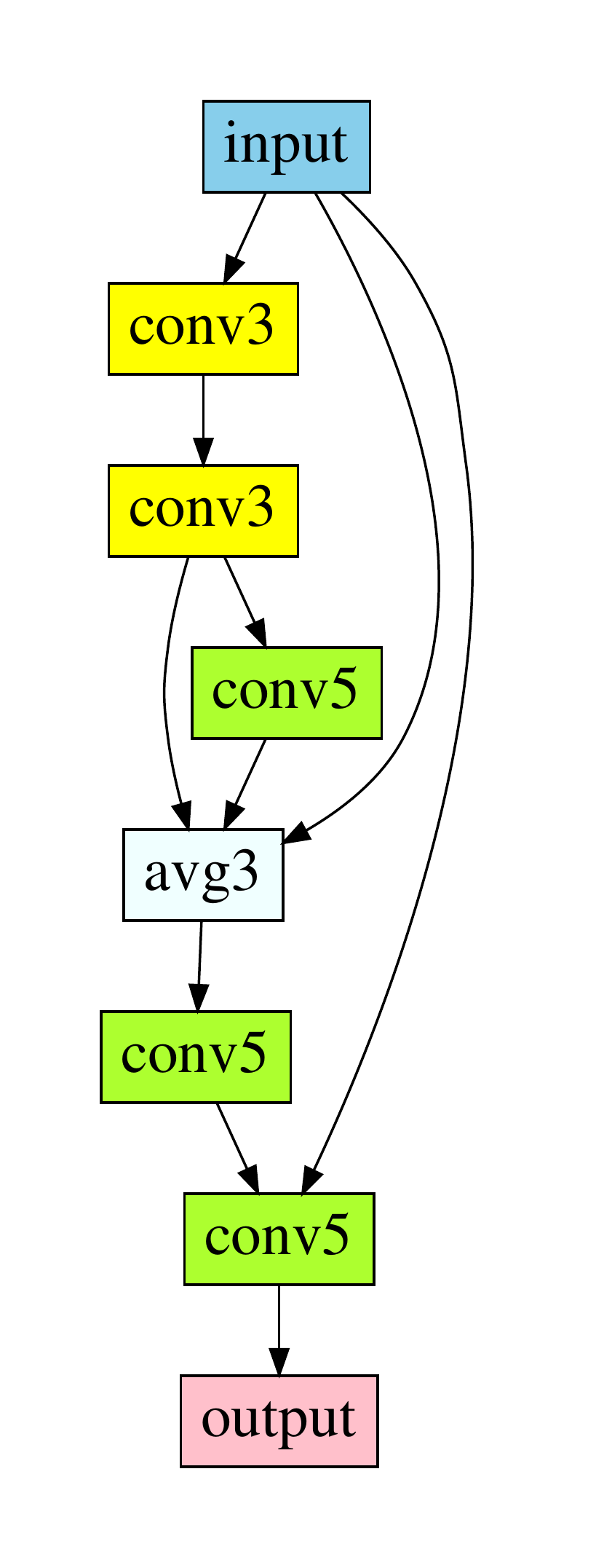}}
\caption{Illustration of the top 5 performing architectures found by \textbf{D-VAE}. The reported test accuracies (from left to right) are: $94.80\% , 94.74\% , 94.70\% , 94.63\%, 94.63$\%.}
\label{exp3:dvae}
\end{figure}

In this experiment, we perform Bayesian optimization in order to generate high-performing architectures, using DVAE-EMB and D-VAE.
Following the methodology adopted by \cite{zhang2019dvae}, we perform 10 iterations of batch Bayesian Optimization and we report the average accuracy results across 10 trials. We use expected improvement (EI)~\cite{10.1007/3-540-07165-2_55}
as the acquisition function. Moreover,  we use the SGP from the previous experiment to model the distribution of the objective function. In every iteration, we select a 50-sample batch by maximizing the acquisition function. For each batch, the selected latent representations are decoded into network architectures and are evaluated using their weight-sharing accuracy on CIFAR-10. The decoded architectures are added to the training set and the SGP is retrained, initiating the next BO iteration. Finally, we select the 5 top-performing generated architectures and we fully train them on CIFAR-10 to evaluate their true performance, following the training procedure of~\cite{pham2018efficient}. \par 
In Figures \ref{exp3:dvae_emb3}, \ref{exp3:dvae}, we visualize the top-5 architectures discovered
by DVAE-EMB and D-VAE  and make the following observations: \begin{enumerate}[font=\bfseries]
  \setlength\itemsep{.2em}
    \vspace{-.2cm}

    \item DVAE-EMB generate better architectures than D-VAE. Our best architecture achieves accuracy equal to $95.33\%$, while D-VAE's highest accuracy is $94.80\%$. This indicates that our proposed method leads to the construction of a very efficient latent space for searching neural network architectures. Moreover, our top-5 architectures achieve accuracy higher than $95\%$, therefore our model is able to learn not only a single high-performing architecture, but a distribution of such architectures.

\item  The top architectures generated from DVAE-EMB present a smoother operation transition than those generated from D-VAE. Specifically, we observe that DVAE's architectures present a diversity of operations, in contrast with DVAE-EMB in which two operations (convolution 5x5 and separable convolution 5x5) are mostly used.
Intuitively, DVAE-EMB learned the computational similarity of those two operations, and encoded the top-performing architectures in similar points on latent space.
\item The top-performing architectures share the same structural patterns. This observation is supported by previous works that highlight the strong effect of the wiring patterns of the layers on the performance of the architecture~\cite{wortsman2019discovering,pmlr-v119-you20b}.
\end{enumerate}  

\subsection{Architecture Performance and Graph Properties }
The common graph structures that we observed in Figures \ref{exp3:dvae_emb3} and \ref{exp3:dvae} lead us to investigate which graph characteristics are the most informative about the performance of the architecture. 
For this reason, we monitored the structural patterns of the 19,020 architectures generated in the ENAS benchmark~\cite{pham2018efficient} and we computed various graph metrics. Two of these metrics, a) the \textit{clustering coefficient} and b) the \textit{average path length} reveal a correlation with the architecture performance. Specifically, we clustered the architectures into six groups according to their performance and we measured their distributions with respect to the examined properties.  

\begin{figure}[t]
    \centering
    \includegraphics[width=0.38\textwidth]{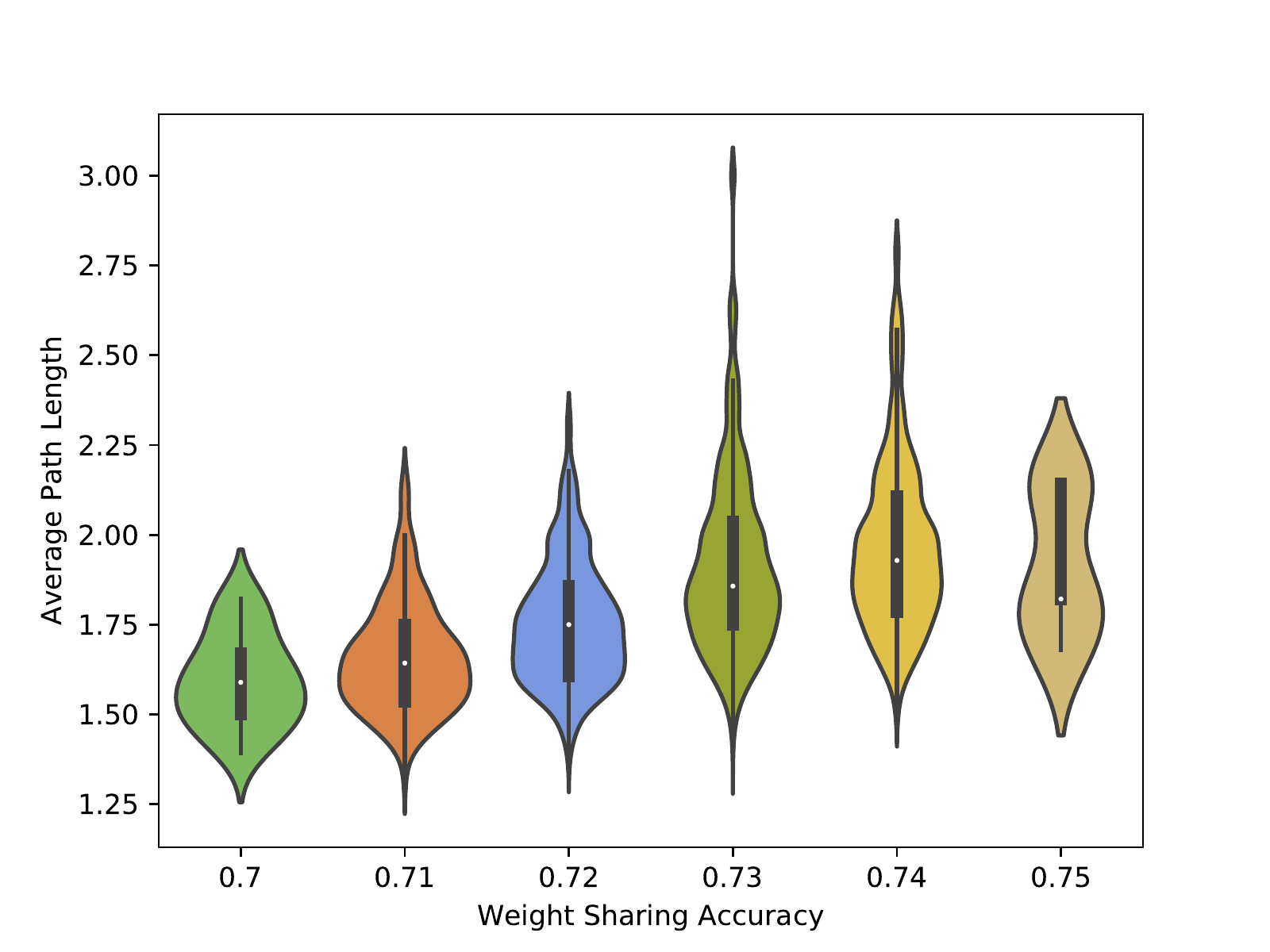}
    \caption{Average path length with respect to model performance. Pearson's $r = 0.32$}
    \label{fig:apl}
\end{figure}

\begin{figure}[t]
    \centering
    \includegraphics[width=0.38\textwidth]{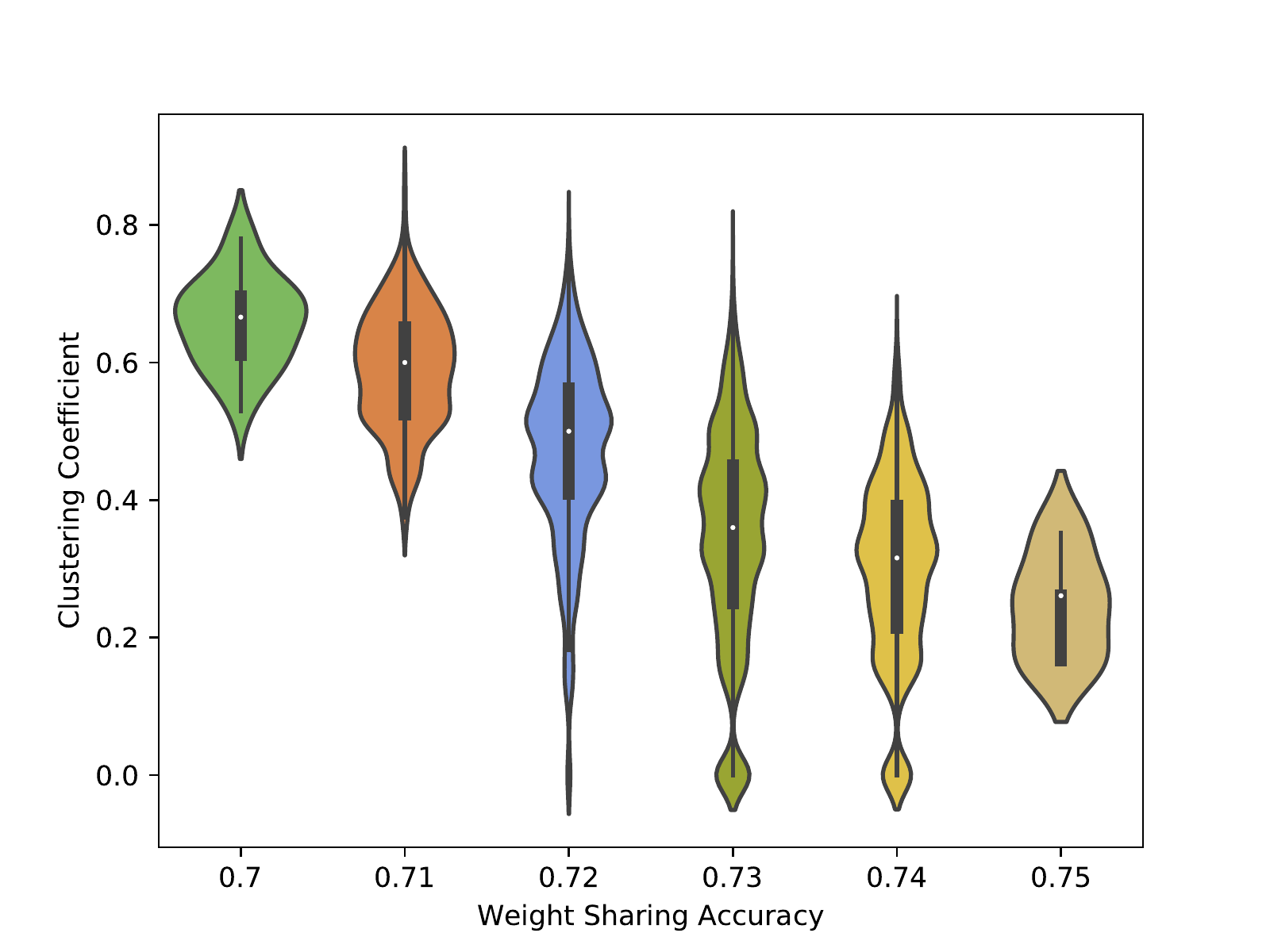}
    \caption{Clustering coefficient with respect to model performance. Pearson's $r = -0.39$}
    \label{fig:cc}
\end{figure}

The results are visualized in Figures \ref{fig:apl} and \ref{fig:cc}. We can observe that the mean average path length is increasing, whereas the clustering coefficient is decreasing, as we move to groups of architectures with higher performance. These findings are aligned with the corresponding Pearson correlation coefficients of the two metrics with the model performance. In particular, Pearson's $r$ between the average path length and the performance is $0.32$ indicating a positive correlation, while Pearson's $r$ between the clustering coefficient and the performance is $- 0.39$ indicating a negative correlation. 


\subsection{Latent Space Visualization}
\begin{figure*}[t]
\centering  
\subfigure[Weight Sharing Accuracy]{\includegraphics[scale=0.7]{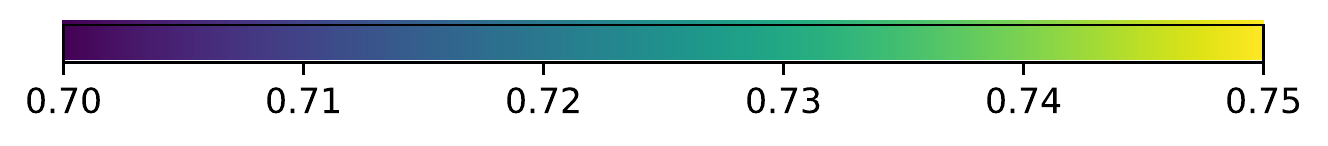}}

\subfigure[]{\label{dvae_a}\includegraphics[width=0.30\linewidth]{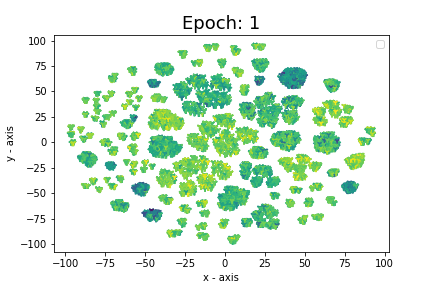}}
\subfigure[]{\includegraphics[width=0.30\linewidth]{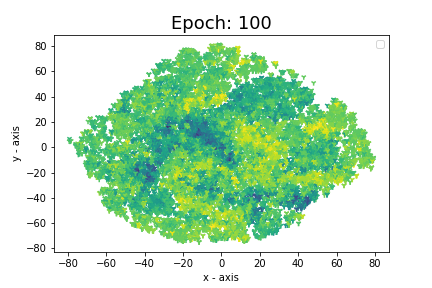}}
\subfigure[]{\includegraphics[width=0.30\linewidth]{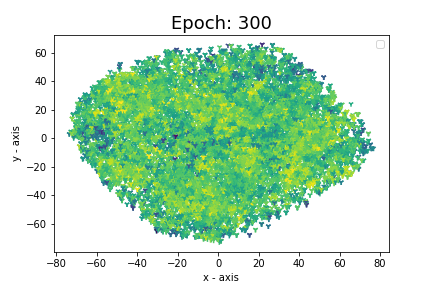}}
\caption{2-D Visualization of the latent space learned by \textbf{D-VAE} during training.}
\label{exp5:DVAE}
\subfigure[]{\label{dvae_emb4_a}\includegraphics[width=0.30\linewidth]{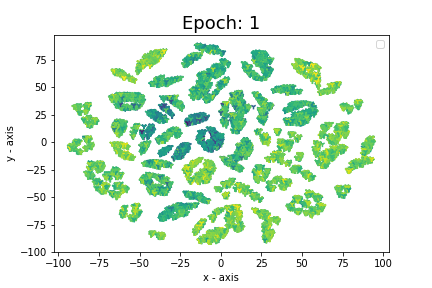}}
\subfigure[]{\includegraphics[width=0.30\linewidth]{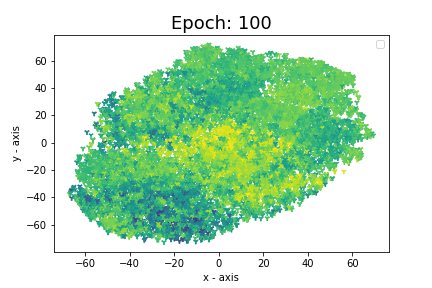}}
\subfigure[]{\label{dvae_emb4_final}\includegraphics[width=0.33\linewidth]{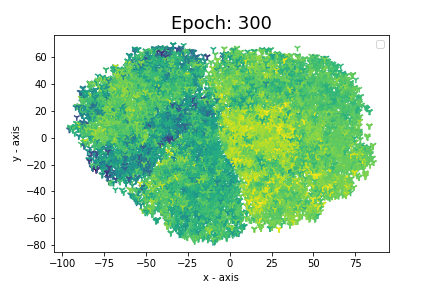}}
\caption{2-D Visualization of the latent space learned by our model \textbf{DVAE-EMB} during training.}
\label{exp5:our_model}
\end{figure*}

In this experiment, we compare the produced latent space of D-VAE with that of DVAE-EMB, projected in 2D space using t-SNE \cite{vanDerMaaten2008}. The visualizations are presented in Figures \ref{exp5:DVAE} and \ref{exp5:our_model}, where the weight sharing accuracy is color encoded. 

In the first epoch (Figures \ref{dvae_a} and \ref{dvae_emb4_a}) the auto-encoder is not yet trained and the representations do not form a continuous latent space. Therefore, multiple architectures are mapped to the same representation making the continuous optimization method not feasible. As the training proceeds, we observe that the architecture representations of both models span the whole latent space. This indicates that the auto-encoders are able to produce a $1-1$ correspondence between the architectures and the latent representations.

Regarding the smoothness of the latent space, DVAE-EMB can accurately cluster together the high-performing architectures, as shown in Figure~\ref{dvae_emb4_final}. This is the most important property in our application, as a smooth latent space can significantly enhance the performance of the search strategy. Note that the latent space was constructed in a fully-unsupervised manner, without having an accuracy signal of the architectures. Therefore, the smoothness is achieved by leveraging only the graph structure and the operation information of the architectures. In contrast, in the D-VAE's latent space the transition of accuracy is not smooth. The high performing architectures are located all over the latent space, without forming clusters. This indicates that our operation embeddings method benefits the process of mapping similar operations together and hence mapping similar performance architectures together. 

\section{Conclusion}\label{sec:conclusion}
Graph-based NAS methods have focused so far on encoding the structural properties of architectures, assuming a fixed representation of the performed operations. In this work, we introduce operation embeddings as a way to replace one-hot encodings of operations with learnable continuous representations that are incorporated into the optimization process. 
Our method enables the NAS framework to learn the computational and structural relationships of different operations, leading to a more accurate architecture latent space. 
The introduced approach has been evaluated on an exhaustive experimental study in ENAS benchmark, highlighting the effectiveness of operation embeddings. Our findings indicate that operation embeddings lead to shorter training time, smoother architecture representations and enhanced performance of various NAS models. We hope that the effectiveness and the flexibility of operation embeddings can motivate future studies to explore the representation power of the operation encoding.

{\small
\bibliographystyle{ieee_fullname}
\bibliography{arxiv}
}

\clearpage
\begin{appendix}
    \renewcommand{\thesection}{A}
   
\section{Model Details} 
In this section we provide more details about the baseline model  D-VAE~\cite{zhang2019dvae}, 
We also further describe the incorporation of our proposed method into the D-VAE, that produces DVAE-EMB model, and finally we prove that DVAE-EMB can injectively encode the computations on DAGs.

\paragraph{D-VAE.}  D-VAE is a graph-based variational autoencoder for Directed
Acyclic Graphs (DAGs). It uses a message-passing graph neural network to encode the graphs using an asynchronous message passing schema, following the topological ordering of the DAG. Firstly, for every node $u$, it aggregates the messages from its neighbors, using an aggregation function $A$ \begin{equation}
    h_u^{in} = A(\{ Cat(h_v,x_v) : (v \rightarrow u) \}) \label{eq:dvae_agg}
\end{equation},
where $x_v$ is the \textbf{one-hot vector} of node $v's$ type, and $Cat$ is a concatenation operation.
Secondly, it update the representation of every node $u$, based on the incoming aggregated message from its neighbors and the one-hot vector $x_u$ of node $u's$ type, \begin{equation}
    h_u = U(h_u^{in},x_u). \label{eq:dvae_update}
\end{equation}
In contrast with simultaneous message passing schemas, D-VAE update the hidden states of the nodes following the topological ordering of the DAG. This asynchronous message passing scheme can effectively encode the computations on DAGs, but the one-hot vector representation of the operations limits the expressivity of the model.

As the goal is to perform optimization in the continuous learned space, the encoder must map different architectures to different representations $h_G$. To achieve this, \textit{the aggregation function $A$ and the update function $U$ must be injective}, as noted in Theorem 2 in \cite{zhang2019dvae}. Also, the encoder should be invariant to node permutations, such that isomorphic graphs, that represent the same architecture, be mapped in the same representation. To achieve this, \textit{the aggregation function must be permutation invariant}, as noted in Theorem 1 in \cite{zhang2019dvae}. To model these two functions, they used a gated sum as an aggregation function: \begin{equation}
    h_u^{in} = \sum_{u \rightarrow v} g(Cat(h_v,x_v)) \odot m(Cat(h_v,x_v)), \label{eq:dvae_aggr_model}
    \end{equation}
    where $m$ is a mapping network and $g$ is a gating network
    and a gated recurrent unit (GRU)\cite{gru} as an update function: \begin{equation} \label{eq:dvae_update_model}
    h_u = GRU(x_u,h_u^{in})
\end{equation}.

\paragraph{DVAE-EMB} The model DVAE-EMB replaces the one-hot vectors in D-VAE, with our proposed operation embeddings approach. The aggregation function \ref{eq:dvae_agg} and the update function \ref{eq:dvae_update} are transformed as follows:
\begin{equation}
    h_u^{in} = A(\{ Cat(h_v,O(x_v)) : (v \rightarrow u) \}) \label{eq:dvae_emb_agg}
\end{equation},

\begin{equation}
    h_u = U(h_u^{in},O(x_u)) \label{eq:dvae_emb_update} ,
\end{equation}
where $O(x_u)$ is the operation embedding of node's $u$ type.  The equations \ref{eq:dvae_aggr_model},\ref{eq:dvae_update_model} are now modified as follows:
\begin{equation}
    h_u^{in} = \sum_{u \rightarrow v} g(Cat(h_v,O(x_v))) 
    \odot m(Cat(h_v,O(x_v))) \label{eq:dvae_emb_aggr_model}
\end{equation}
    \begin{equation} \label{eq:dvae_emb_update_model}
    h_u = GRU(O(x_u),h_u^{in})
\end{equation}.

Since our operation embedding function $O$ is injective, the update and the aggregation functions remain injective, as the composition of injective functions is injective. Therefore Theorems 1,2 still holds for DVAE-EMB and consequently our encoder can injectively encode the computations on DAGs. 





\renewcommand{\thesection}{B}
\section{Training Details} In order to have a fair comparison, we use the same settings from Zhang et al.~\cite{zhang2019dvae} to train our models (DVAE-EMB, GCN-EMB). For the baselines models, we use the reported results from Zhang et al.~\cite{zhang2019dvae}. We set the dimensionality of the operation embeddings to be $3$ for both models. 

For DVAE-EMB we employ the strategy described in Section 3.3. Specifically, we fully-train the model
for 4 iterations, for 300 epochs in each iteration. In the first iteration we initialize the operation embeddings from a normal distribution $\mathcal{N}(\textbf{0},\textbf{I})$. In the next iterations, we initialize the operation embeddings, using the output of the last epoch in the previous iteration. Using this strategy, we observe an increasing performance of the autoencoder, as the operation embeddings are trained for more epochs and capture more effectively the relations between the operations.   

For GCN-EMB we just initialize the operation embeddings from a normal distribution, and train the model for 300 epochs without extra iterations. Note that both DVAE-EMB and GCN-EMB achieve better results from their counterparts (DVAE,GCN) even from the first iteration of the operation embeddings. 

All models are implemented using Pytorch library~\cite{NEURIPS2019_9015}. The code is available at \url{https://github.com/MichailChatzianastasis/Graph-based_NAS_with_Operation_Embeddings}.

\renewcommand{\thesection}{C}
\section{Architecture Performance and Graph Properties - Experiment Details}

In this experiment, we investigate the relation between the performance of the architecture and its corresponding graph structure as described in Section 3.1.  Specifically, two graph properties, the \textit{average path length} and the \textit{clustering coefficient} reveal a correlation with the architecture performance. 

Average path length is defined as the average shortest path distance between all possible pairs of network nodes. We calculate average path length of a graph $G$ using the following formula:
\begin{equation}
L_G = \dfrac{1}{n \cdot (n-1)} \cdot \sum_{i\neq j} d(u_i,u_j),
\end{equation}
where $d(u_i,u_j)$ is the shortest distance between the nodes $u_i$ and $u_j$. Assume that $ d(u_{1},u_{2})=0$ if there is no path from $u_i$ to $u_j$. 

Clustering coefficient measures the probability that the adjacent vertices of a vertex are connected. We calculate this property using the igraph software \cite{igraph}. Specifically, we measure the global clustering coefficient which is the the ratio of the triangles and the connected triples in the graph. Because we have directed graphs the direction of the edges is ignored.
\end{appendix}

\end{document}